\documentclass[10pt,twocolumn,letterpaper]{article}

\usepackage{cvpr}      

\usepackage{tikz}
\usetikzlibrary{positioning, fit} 
\usepackage{caption}

\definecolor{cvprblue}{rgb}{0.21,0.49,0.74}
\usepackage[pagebackref,breaklinks,colorlinks,allcolors=cvprblue]{hyperref}

\def\paperID{*****} 
\def\confName{CVPR}
\def\confYear{2026}

\title{ A Non-Invasive Alternative to RFID: Self-Sufficient 3D Identification of Group-Housed Livestock}

\author{
Shiva Paudel \quad TsungCheng Tsai \quad Dongyi Wang \\
University of Arkansas \\
Fayetteville, Arkansas USA \\
{\tt\small shivap@uark.edu \quad ttsai@uark.edu \quad dongyiw@uark.edu}
}

\begin{document}
\makeatletter
\apptocmd{\@maketitle}{
   \centering
   \vspace{-5mm} 
   \includegraphics[width=\linewidth]{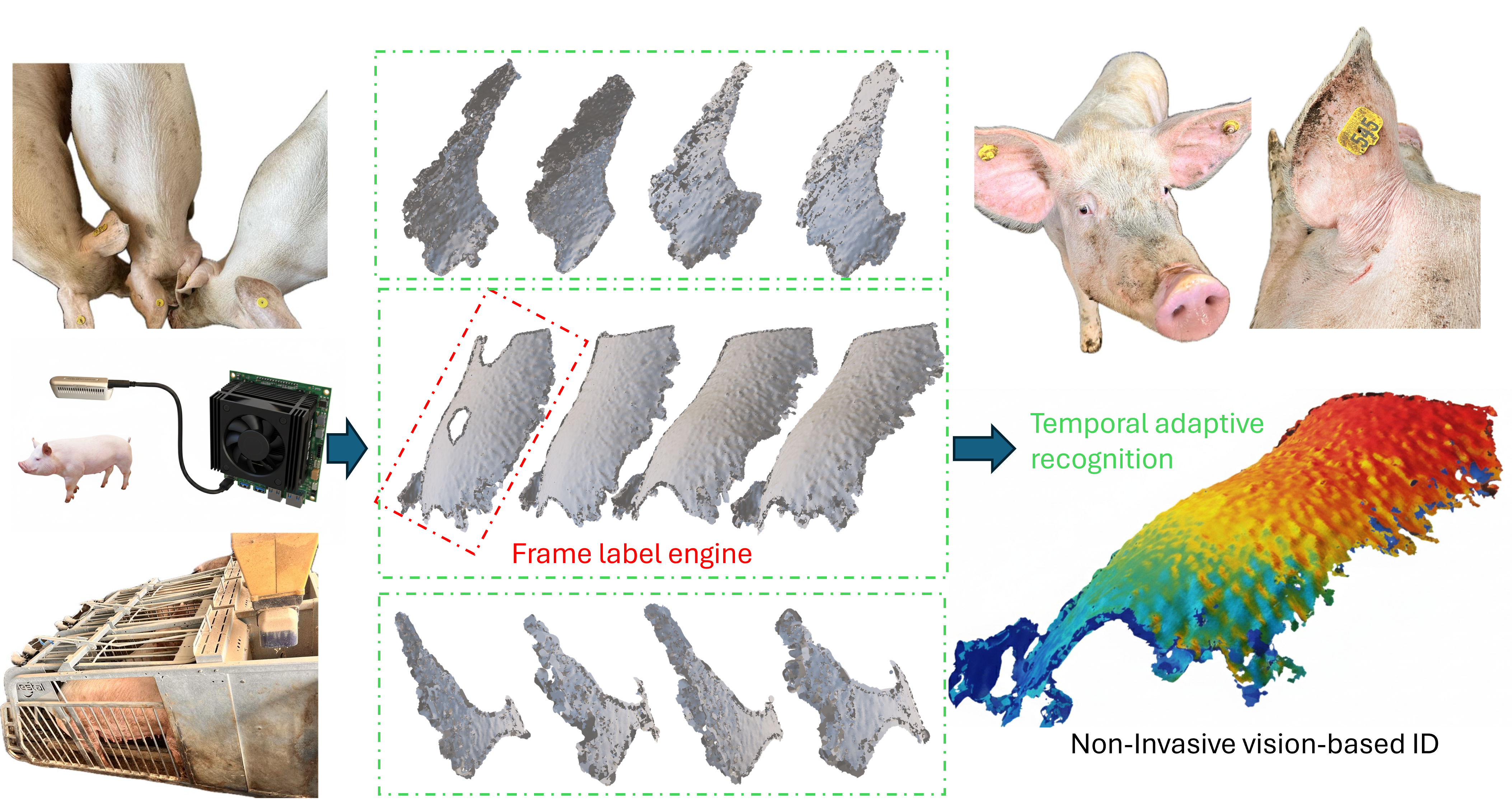}
   \captionof{figure}{Conceptual framework for non-invasive, vision-based livestock identification. Moving beyond costly and invasive RFID tagging, our approach treats identification as a temporal consensus problem. By integrating frame-level deep learning with a visit-level consensus mechanism, the system filters transient noise to assign a single, robust ID per visit. This spatiotemporal aggregation achieves 100\% identification accuracy on group-housed sow point clouds.}
   \label{fig:setup}
   \vspace{8mm} 
}{}{}
\makeatother

\maketitle
\begin{abstract}
Accurate identification of individual farm animals in group-housed environments is a cornerstone of precision livestock management. However, current industry standards rely heavily on Radio Frequency Identification (RFID) ear tags, which are invasive, prone to loss, and restricted by the spatial limitations of antenna fields. In this paper, we propose a non-intrusive, vision-based identification system leveraging 3D point cloud data captured within a commercial electronic feeding station (EFS). Departing from traditional supervised frame-level inference, we introduce the Temporal Adaptive Recognition Architecture (TARA), a self-sufficient, semi-supervised framework designed to maintain identity consistency over time. TARA employs a dynamic recalibration mechanism that updates individual identity profiles to account for morphological changes in the livestock. To facilitate training in label-scarce environments, we utilize a visit-level majority voting strategy to generate high-fidelity pseudo-labels from raw temporal sequences. Experimental results on a group housed sow dataset collected from an operational commercial barn demonstrate that our approach achieves 100\% \ identification accuracy at the visit level. These results suggest that vision-based 3D point cloud analysis offers a robust, superior alternative to RFID-based systems, paving the way for fully autonomous individual animal monitoring.  
\end{abstract}    
\section{Introduction}
\label{sec:intro}
Modern livestock production is shifting from group-level optimization toward "individualized precision care," a transition driven by the dual imperatives of enhancing animal welfare and navigating narrow economic margins \cite{rosa2021grand, DBerkman}. This paradigm requires monitoring each animal as a distinct biological entity to facilitate early health interventions, track physiological states, and mitigate social stressors. By prioritizing the individual, producers move from reactive, herd-wide management to proactive care that improves both longevity and quality of life. However, industry-standard RFID systems remain suboptimal, as invasive ear-tagging induces acute stress and requires significant maintenance labor \cite{silva2006evaluating, jang2022management, paudel2025advancements}. This technological bottleneck is especially critical in managing gestating sows, where non-invasive, continuous identification is the missing link required to bridge the gap between industrial housing and the high-precision care necessary for maternal health and reproductive success.

To address these limitations, we propose the Temporal Adaptive Recognition Architecture (TARA), a non-invasive framework for identifying group-housed sows using 3D point cloud sequences. Unlike 2D imagery, which is susceptible to volatile lighting, TARA leverages 3D geometric consistency to differentiate individuals with similar morphologies by treating feeding events as discrete "visits" and aggregating frame-level data into a session-level consensus. Our primary contributions include a robust 3D vision pipeline tailored for commercial facilities and a Temporal Consensus Mechanism that ensures stable identification across entire visits. Furthermore, we introduce an Autonomous Re-calibration Loop that uses high-confidence signatures as pseudo-labels, allowing the model to adapt to long-term morphological changes such as growth and pregnancy, without requiring human intervention or traditional RFID reliance.

The rest of this paper is structured as follows: Section \ref{sec:related work} reviews 3D deep learning in livestock monitoring, and Section \ref{sec:methods} details the TARA architecture. Section \ref{sec:res_and_dis} present experimental results and the autonomous self-training loop's efficacy. Finally, Section \ref{sec:conclusion} summarizes our findings and future implications for vision-based identification.

\section{Related Work}
\label{sec:related work}

Vision-based individual identification is a cornerstone of precision livestock farming (PLF), yet it remains a persistent challenge due to the harsh conditions of commercial barns \cite{kleen2023precision}. Early strategies relied on invasive physical identifiers like ear tags \cite{paudel2025advancements}. While research shifted toward non-contact biometric identification using 2D RGB data, these systems are frequently hampered by feeding-related occlusions and a fundamental inability to distinguish between monochromatic breeds with uniform coat textures, such as white swine or black cattle \cite{hansen2018towards, sharma2025universal}. Consequently, the field has pivoted toward 3D modalities that capture an animal’s inherent geometric structure. While architectures like PointNet and PointMLP have improved recognition, these spatial-only paradigms treat point clouds as isolated data points, making them highly sensitive to postural noise and sensor artifacts. This lack of temporal logic often results in identification "flickering," leaving a critical gap in maintaining identity consistency over time \cite{li2026dk, liang2024study}.

The final frontier in livestock monitoring involves managing the non-stationary nature of animal data, where rapid growth and pregnancy-related changes cause model performance to decay within a "retraining window" of just five to seven days \cite{paudel2025deep}. Current adaptation frameworks and state-space models offer mathematical foundations for online updates but often lack the domain-specific "session consensus" required for noisy, unconstrained farm environments \cite{boudiaf2022parameter, schirmer2024temporal}. Our TARA framework addresses these gaps by leveraging the naturally recurring structure of feeding visits to generate session-level pseudo-labels. This autonomous re-calibration loop provides a self-supervised solution to long-term morphological drift, ensuring longitudinal identification stability without the need for frequent human intervention or the limitations of traditional open-world formulations \cite{Joseph_2021_CVPR}.

\section{Methods}
\label{sec:methods}

The image acquisition platform was deployed at the University of Arkansas Animal Research Facility, USA (IACUC protocol \#25059).

\subsection{Problem Setup and Notation}
We observe a continuous stream of depth frames and transform each into a 3D point cloud $x \in \mathbb{R}^{N \times 3}$. The stream is naturally partitioned into \textit{visits} (sessions) $v$, where a visit $V_v=\{x_{v,1},\dots,x_{v,N_v}\}$ represents a temporally contiguous sequence of $N_v$ frames captured while an animal is present at a feeding station. We consider a stable population of $C$ individuals. 

The task is to learn a frame-level classifier $f_\theta: \mathbb{R}^{N \times 3} \rightarrow [0,1]^C$ that outputs a probability distribution $\mathbf{p}_{v,i}=f_\theta(x_{v,i})$. From these, we must infer a robust visit-level identity $\widehat{Y}_v$ and periodically update the parameters $\theta$ to accommodate the non-stationary nature of livestock morphology (e.g., growth and pregnancy) using high-confidence sessions as self-supervised pseudo-labels.

\subsection{TARA: Temporal Adaptive Recognition Architecture}
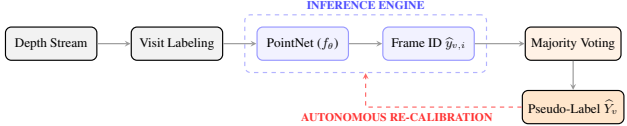
\begin{figure}[ht]
    \centering
    \resizebox{\columnwidth}{!}{
    \begin{tikzpicture}[
        node distance=0.8cm and 0.8cm,
        box/.style={rectangle, draw, thick, rounded corners, text centered, minimum height=0.8cm, minimum width=2.2cm, font=\small, fill=white, inner sep=4pt, align=center},
        innerbox/.style={fill=blue!5, draw=blue!40, thick},
        arrow/.style={thick, ->, >=stealth, draw=gray!80},
    ]
    \node (stream) [box, fill=gray!10] {Depth Stream};
    \node (group) [box, fill=gray!10, right=of stream] {Visit Labeling};

    \begin{scope}[local bounding box=frameengine]
        \node (pointnet) [box, innerbox, right=of group] {PointNet ($f_\theta$)};
        \node (rawid) [box, innerbox, right=of pointnet] {Frame ID $\widehat{y}_{v,i}$};
    \end{scope}

    \node[draw, blue!30, thick, dashed, rounded corners, fit=(pointnet) (rawid), inner sep=8pt] (enginebox) {};
    \node[anchor=south, blue!70, font=\footnotesize\bfseries, yshift=0pt] at (enginebox.north) {INFERENCE ENGINE};

    \node (mv) [box, fill=orange!10, right=1.2cm of rawid] {Majority Voting};
    \node (pseudo) [box, fill=orange!20, below=0.7cm of mv] {Pseudo-Label $\widehat{Y}_v$};

    \draw [arrow] (stream) -- (group);
    \draw [arrow] (group) -- (pointnet);
    \draw [arrow] (pointnet) -- (rawid);
    \draw [arrow] (rawid) -- (mv);
    \draw [arrow] (mv) -- (pseudo);

    \draw [arrow, thick, red!60, dashed] (pseudo.west) 
        -| (enginebox.south) 
        node[pos=0.4, below, font=\footnotesize\bfseries, color=red!80] {AUTONOMOUS RE-CALIBRATION};
    \end{tikzpicture}
    }
    \caption{TARA framework: Depth stream grouping, frame inference, and visit consensus with self-supervised feedback.}
    \label{fig:tara_compact}
\end{figure}

The TARA framework (Fig.\ref{fig:tara_compact}) introduces a hierarchical approach to identification, moving beyond isolated frame analysis to exploit the temporal stability of a feeding visit.

\subsubsection{Temporal Majority Voting (Visit Consensus)}
TARA assumes that an animal's identity remains invariant throughout the duration of a feeding visit. During inference, we aggregate all frame-level predictions $\widehat{y}_{v,i}$ within a visit. By applying a majority voting scheme, the system becomes resilient to transient sensor artifacts and "postural noise" (e.g., a sow turning its head or moving out of the optimal sensing volume), provided the majority of frames yield the correct classification.

\subsubsection{Autonomous Re-calibration via Pseudo-Labels}
A critical challenge in precision livestock farming (PLF) is model decay; as animals grow, their 3D signatures drift away from the training distribution. TARA addresses this via a self-supervised feedback loop. Visits that exhibit a high "Consensus Strength" ($\rho_v \ge \gamma$) are deemed highly reliable. These visits are automatically retroactively labeled with the predicted visit identity $\widehat{Y}_v$ and added to a dynamic fine-tuning pool. This allows the PointNet backbone to adapt to gradual morphological changes without requiring human-labeled ground truth or a constant RFID tether.

\subsection{Mathematical Framework for TARA}
TARA's logic is defined by confidence filtering and temporal aggregation.

\subsubsection{Frame-Level Inference and Confidence Filtering}
For every visit $v$, each frame $i$ produces a probability vector $\mathbf{p}_{v,i} \in [0,1]^C$. To ensure the integrity of the re-calibration loop, we apply a strict confidence threshold $\tau=0.99$. We define the set of valid frames $\mathcal{J}_v$ as:
\[
\mathcal{J}_v = \{ i : \max_c p_{v,i}(c) \ge \tau \}, \quad M_v = |\mathcal{J}_v|.
\]
where $M_v$ is the count of valid frames. For $i\in\mathcal{J}_v$, the predicted frame-level class is $\widehat{y}_{v,i}=\arg\max_c p_{v,i}(c)$.

\subsubsection{Visit-Level Majority Consensus}
The frequency of each predicted class within a visit is calculated as:
\[
n_v(c)=\sum_{i\in\mathcal{J}_v}\mathbf{1}[\widehat{y}_{v,i}=c],
\]
where $\mathbf{1}[\cdot]$ is the indicator function. The majority class $c_v^*$ and the consensus strength $\rho_v$ are:
\[
c_v^*=\arg\max_c n_v(c), \quad \rho_v=\frac{ n_v(c_v^*)}{M_v}.
\]
To avoid "ID flickering" and ensure longitudinal stability, a final identity $\widehat{Y}_v$ is assigned only if the visit meets minimum duration ($K=10$ frames) and consensus ($\gamma=0.50$) requirements:
\[
\widehat{Y}_v=
\begin{cases}
c_v^*, & \text{if } M_v \ge K \text{ and } \rho_v \ge \gamma,\\
\emptyset, & \text{otherwise.}
\end{cases}
\]

\subsection{PointNet Backbone and Spatial Transformations}
The primary feature extractor is a PointNet architecture \cite{qi2017pointnet}, selected for its permutation invariance and inherent ability to process unordered point sets. To achieve robustness against postural variance, the framework utilizes dual "T-Net" modules. First, an Input Transform mini-network predicts a $3 \times 3$ affine transformation matrix to align raw point clouds into a canonical space. This is followed by a Feature Transform, a second $64 \times 64$ transformer that aligns high-dimensional features, ensuring the learned signatures remain invariant to the subject's orientation relative to the camera. The model is trained as a $C$-class classifier optimized via categorical cross-entropy loss. To establish a "pre-drift" baseline, the architecture is trained on a 80/20 train-validation split using data exclusively curated from Day 1.

\subsection{Data Acquisition and Preprocessing}
The system was deployed at an animal research facility utilizing three Intel RealSense D435 sensors mounted on Gestal ESF feeding stations (Fig. \ref{fig:setup}, where high-traffic data capture was triggered at $\Delta t = 2$ s only when subjects were within 0.6 m, yielding 89,944 RGB-D pairs over a 96-hour period. To ensure ground-truth alignment, vision visits were segmented by inter-frame intervals of less than 5 s and synchronized with Gestal RFID logs; IDs were assigned only if a vision visit was strictly contained within a single RFID record's duration, with overlapping or missing entries discarded to maintain label integrity. Subsequent point cloud processing involved 5 mm voxel downsampling to normalize density, followed by a Region of Interest (ROI) crop between 0.1 m and 0.6 m and connectivity analysis to isolate the dorsal surface from stall structures. Finally, inputs were centered at the origin and normalized to a unit sphere, ensuring that subsequent feature extraction remained focused on the subjects' distinct morphology.

\begin{figure}[htbp]
    \centering
    \begin{minipage}[b]{0.48\columnwidth}
        \centering
        \begin{subfigure}{\textwidth}
            \centering
            \includegraphics[width=\linewidth, height=2.5cm, keepaspectratio]{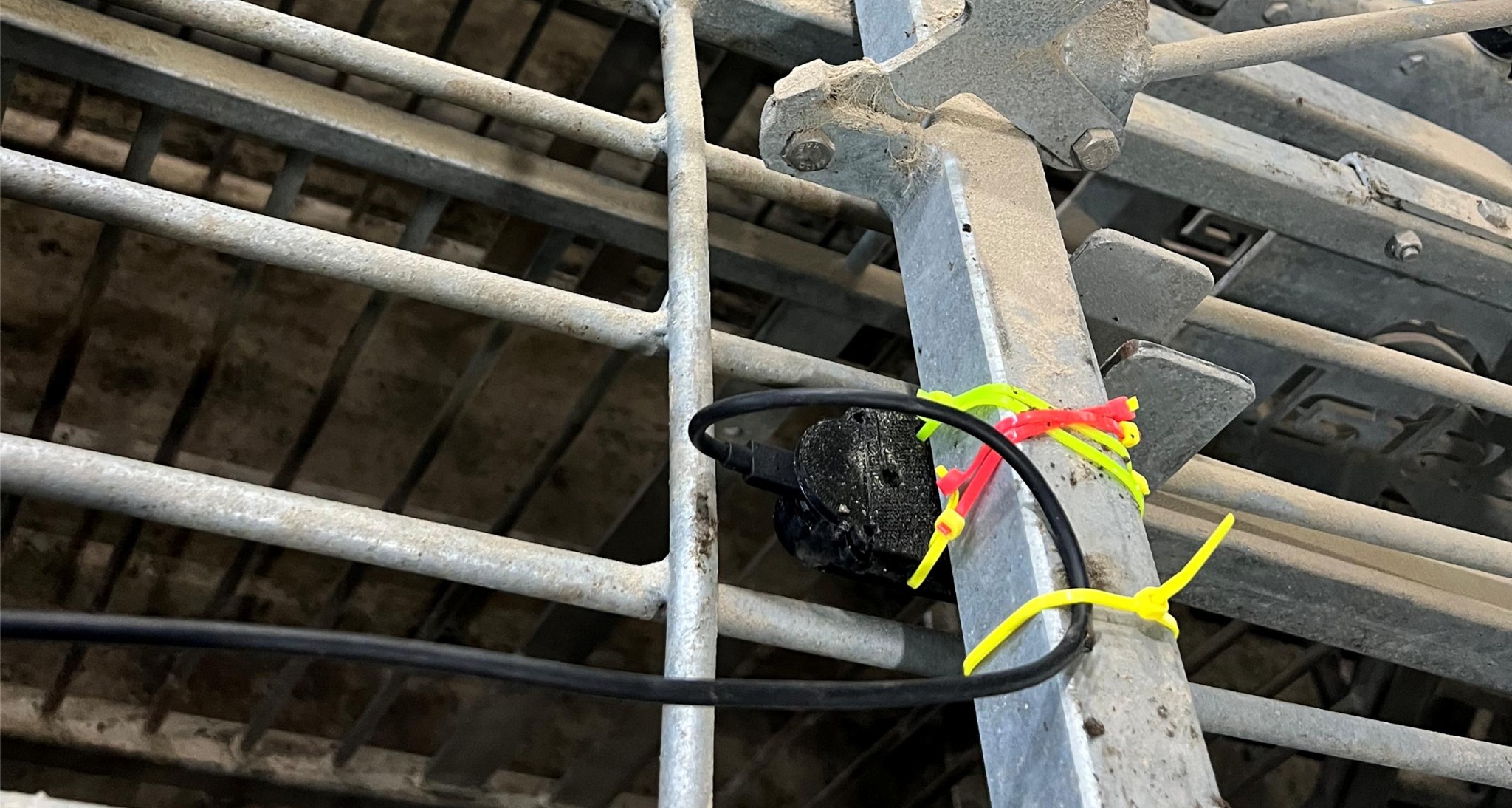}
            \caption{} \label{fig:top_view}
        \end{subfigure}
        \vspace{3pt}
        \begin{subfigure}{0.48\textwidth}
            \centering
            \includegraphics[width=\linewidth, height=1.75cm, keepaspectratio]{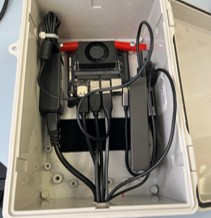}
            \caption{} \label{fig:internal}
        \end{subfigure}
        \hfill
        \begin{subfigure}{0.48\textwidth}
            \centering
            \includegraphics[width=\linewidth, height=1.75cm, keepaspectratio]{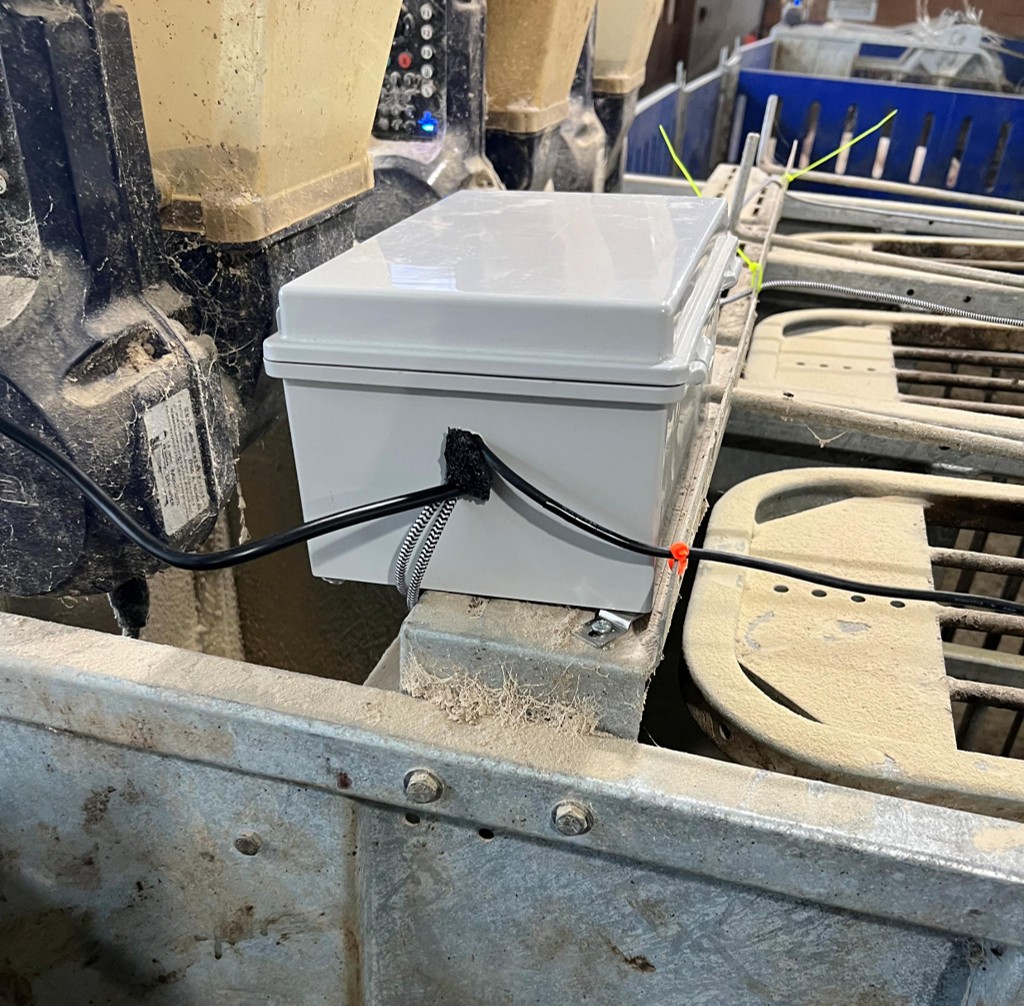}
            \caption{} \label{fig:housing}
        \end{subfigure}
    \end{minipage}
    \hfill
    \begin{minipage}[b]{0.48\columnwidth}
        \centering
        \begin{subfigure}{\textwidth}
            \centering
            \includegraphics[width=\linewidth, height=4.55cm]{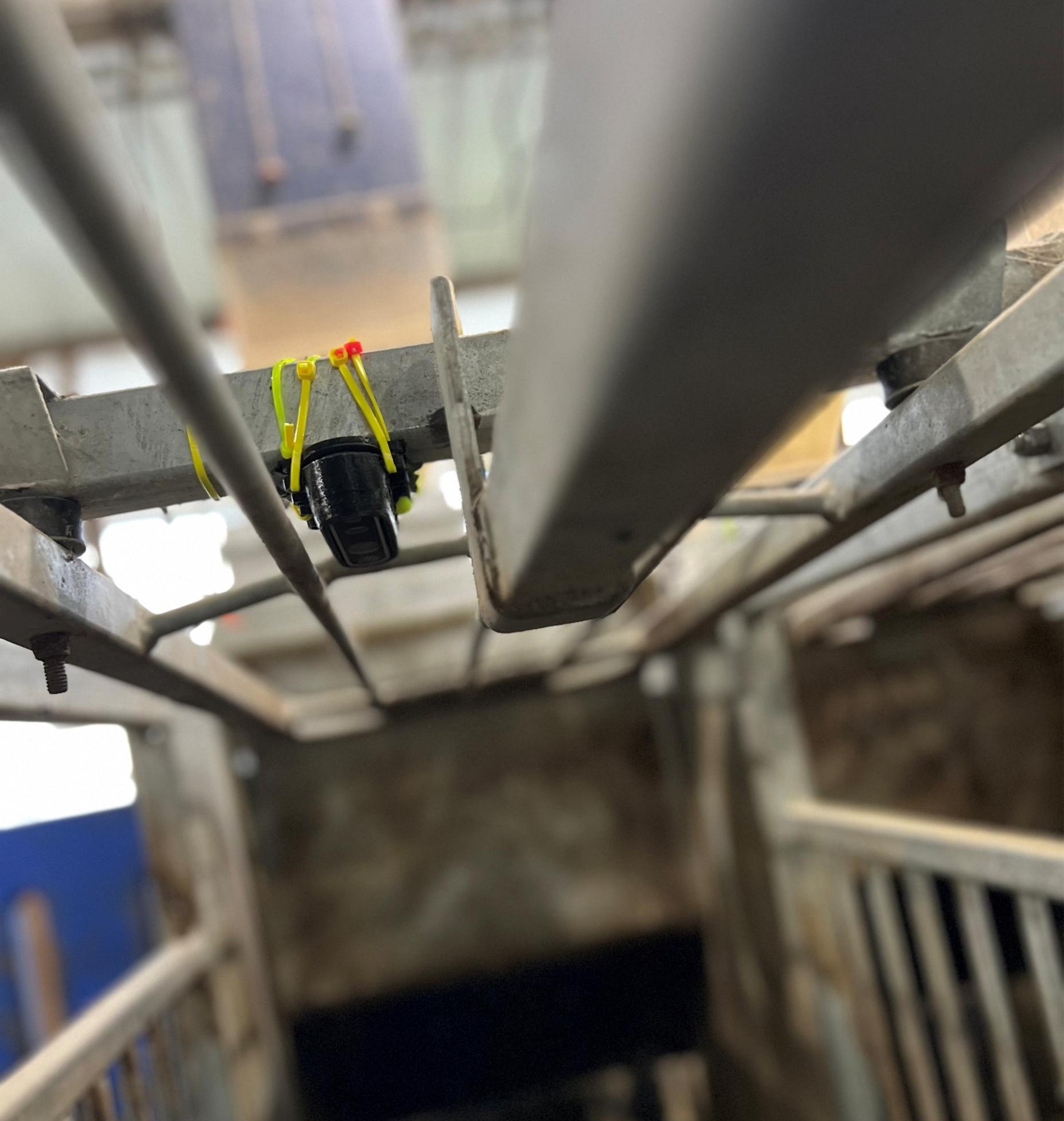}
            \caption{} \label{fig:wide_view}
        \end{subfigure}
    \end{minipage}
    \caption{Experimental setup: \subref{fig:top_view} top view, \subref{fig:internal} edge unit, \subref{fig:housing} housing, and \subref{fig:wide_view} wide view. Intel RealSense D435 cameras monitor 19 group-housed sows via Jetson Orin Nano edge modules.}
    \label{fig:setup}
\end{figure}

\subsection{Evaluation Metrics}
We assess TARA across two hierarchical levels:
\begin{equation}
\eta_{\text{fr}} = \frac{1}{M} \sum_{(v,i)\in\mathcal{F}} \mathbf{1}[\widehat{y}{v,i}=y_v], \quad \eta{\text{vis}} = \frac{1}{|\mathcal{V}|} \sum_{v\in\mathcal{V}} \mathbf{1}[\widehat{Y}_v = y_v]
\end{equation}
representing the percentage of correctly identified high-confidence frames ($\tau \ge 0.99$) and valid consensus-filtered sessions ($\rho_v \ge \gamma$), respectively.

\section{Results and Discussion}
\label{sec:res_and_dis}

\noindent \textbf{PointNet Baseline \& TARA Evaluation:} 
Our inference engine utilizes dual T-nets and orthogonal regularization to extract global signatures from 1,500-point dorsal manifolds. Tested on a longitudinal dataset of 9 sows (89,944 frames), we evaluate TARA's ability to mitigate morphological drift across four 24-hour blocks. As shown in Table \ref{tab:combined_results}, the "Base Model" trained on Day 1 shows immediate decay by Day 3. However, the TARA consensus module stabilizes frame-level noise, elevating visit accuracy from 96.30\% to 100\% through iterative updates.

\begin{table}[ht]
\centering
\caption{System Performance: Baseline vs. Iterative TARA Re-calibration.}
\label{tab:combined_results}
\resizebox{\columnwidth}{!}{%
\begin{tabular}{llccc}
\toprule
\textbf{Model State} & \textbf{Test Set} & \textbf{Frame \%} & \textbf{Visit \%} & \textbf{Conv. \%} \\ \midrule
\textbf{Base Model}  & Day 2             & 93.94             & 96.94             & 82.78             \\
(Day 1 Training)     & Day 3             & 94.17             & 96.30             & 77.14             \\ \midrule
\textbf{Re-calib. 1} & Day 3             & 96.77             & 99.13             & 80.42             \\
(Pseudo Day 2)       & Day 4$^{\dagger}$ & 97.03             & 100.00            & 68.00             \\ \midrule
\textbf{Re-calib. 2} & Day 4$^{\dagger}$ & \textbf{97.73}    & \textbf{100.00}   & \textbf{79.54}    \\ \bottomrule
\end{tabular}}
\vspace{2pt} \\ \scriptsize $^{\dagger}$ \textit{Reduced cohort due to session termination.}
\end{table}

\noindent \textbf{Autonomous Re-calibration Efficacy:} 
The iterative re-calibration stages demonstrate the system's "self-centering" capability. Using high-confidence Day 2 visits as pseudo-labels (Re-calib. 1) improved Day 3 frame accuracy by +2.6\% without human intervention. By the second iteration, the model achieved a perfect 100.00\% visit-level identification rate, effectively bridging the gap between static lab training and dynamic, unconstrained animal growth.

\noindent \textbf{Discussion of Key Findings:}
\begin{itemize}[leftmargin=*, noitemsep, topsep=2pt]
    \item \textbf{Sensing Robustness:} Close-proximity mounting caused a 19.7\% loss of raw frames due to depth clipping. However, TARA's visit-level consensus compensated for this, maintaining accuracy despite hardware-level sensing constraints in cramped industrial environments.
    \item \textbf{Environmental Invariance:} 3D geometric point clouds provided superior invariance to high-contrast shadows and specular reflections compared to traditional RGB systems, which often fail under the dynamic "open-barn" lighting of commercial facilities.
    \item \textbf{Identity Contamination:} To prevent noise from "impatient" animal switching at feeders (transitions $<2$s), TARA's consensus threshold ($\rho_v \ge \gamma$) successfully filters out high-noise sessions, preserving the integrity of the autonomous loop.
\end{itemize}

\noindent \textbf{Practical Implications:} 
By leveraging temporal consistency, this approach achieves "temporal sufficiency" without the labor or stress of passive RFID, restoring visit conversion rates to 79.54\% by Day 4. While these results confirm the stability required for continuous profiling, further validation on larger, higher-density cohorts is necessary to ensure scalability in commercial environments.
\section{Conclusion and Future Work}
\label{sec:conclusion}

This study demonstrates the feasibility of a 3D computer-vision system for identifying group-housed livestock using dorsal point clouds. Our TARA framework achieved a peak frame-level accuracy of 97.73\% and a 100\% visit-level identification rate, confirming that 3D spatial features provide sufficient discriminative power despite phenotypic similarity. While longitudinal analysis revealed performance decay due to morphological drift, the implementation of an autonomous re-calibration loop successfully mitigated this by leveraging pseudo-labeled data to restore visit-to-ID conversion rates from 77.14\% to 80.42\% on Day 3. Furthermore, the system proved robust to environmental challenges such as depth-sensor clipping and non-stationary lighting through a high-agreement consensus mechanism ($\rho_v \ge \gamma$) that filters identity contamination. These results suggest that a self-sustaining, semi-supervised vision system offers a viable, non-intrusive alternative to traditional RFID infrastructure for long-term livestock monitoring in commercial gestation facilities.
{
    \small
    \bibliographystyle{ieeenat_fullname}
    \bibliography{main}
}


\end{document}